# Adaptive Deconvolution-based stereo matching Net for Local Stereo Matching

Xin Ma[1,3*], Zhicheng Zhang[1], Danfeng Wang[1], Yu Luo[2*] and Hui Yuan[4], (SeniorMember, IEEE)

1 School of Information Science and Engineering, Shandong University,Qingdao, China

2College of Geomatics, Shandong University of Science and Technology, Qingdao, China

3 Shenzhen Research Institute, Shandong University, Shenzhen, China

4School of Control Science and Engineering, Shandong University, Jinan 250061, China

*Corresponding author: Xin Ma (max@sdu.edu.cn) and Yu Luo (luoyu@sdust.edu.cn).

## Abstract

In deep learning-based local stereo matching methods, larger image patches usually bring better stereo matching accuracy. However, it is unrealistic to increase the size of the image patch size without restriction. Arbitrarily extending the patch size will change the local stereo matching method into the global stereo matching method, and the matching accuracy will be saturated. We simplified the existing Siamese convolutional network by reducing the number of network parameters and propose an efficient CNN based structure, namely Adaptive Deconvolution-based disparity matching Net (ADSM net) by adding deconvolution layers to learn how to enlarge the size of input feature map for the following convolution layers. Experimental results on the KITTI 2012 and 2015 datasets demonstrate that the proposed method can achieve a good trade-off between accuracy and complexity.

Key words: stereo matching, deep learning, convolutional neural network, KITTI2015, KITTI2012

## 1.Introduction

Three-dimensional scene reconstruction is important in autonomous driving navigation, virtual/augmented reality, etc. The texture/color information of a scene can be captured directly by cameras, whereas the scene geometry (especially the depth information) cannot be obtained easily. There are two ways to obtain the depth information, one is a direct measurement, another is binocular vision. Special devices, like laser detectors and millimeter-wave radars, are needed for direct measurement. Besides the direct measurement, depth information can also be obtained by binocular vision-based stereo matching indirectly. With the great development of high-performance computing devices, the depth information will be estimated easily by using the captured left and right view images without additional expensive devices.

Fig. 1 shows a typical binocular camera model [1]. The two-camera centers are placed on the same horizontal line with a baseline of $B$. Usually, the center of the left camera is set to be the base point, that is left focus. The optical axes of the two cameras are parallel, and the optical axis of the left camera is denoted as the axis $Z$. The image plane of the two cameras is parallel with the $XY$ plane which is perpendicular to the axis $Z$. Suppose a point $P(x, y, z)$ in the 3D space, it can be projected onto the pixel position $p_l$ and $p_r$ of the image plane. The discrepancy between $p_l$ and $p_r$ is defined as the disparity, i.e., $d = |p_l - p_r|$. By using the principle of similar triangles, the depth $z$ of the point $P(x, y, z)$ can be calculated by

$$z = B \times f/d, \quad (1)$$

where $f$ represents the focal length of the camera. The value of $x$ and $y$ can be calculated by

$$x = z \times \frac{x_l}{f} \quad y = z \times \frac{y_l}{f}, \quad (2)$$

where $x_l$ and $y_l$ are the horizontal and vertical coordinates of the point on the left image plane. Because the baseline $B$ and the focal length $f$ can be obtained after camera calibration [2], only the disparity should be calculated to obtain the depth information of a scene.

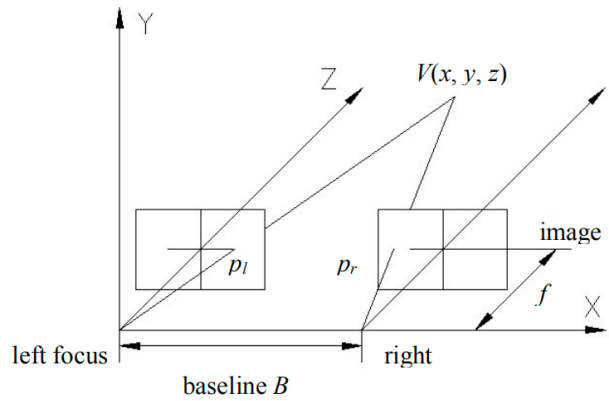

Figure 1. Binocular vision camera model converts binocular vision into a coordinate system.

Stereo matching [3] between the left and the right images can be used to calculate the disparity. Generally speaking, there are three kinds of stereo matching algorithms, i.e., local minimization-based matching, regional minimization-based matching, and global minimization-based matching. With a given matching error function, the local minimization-based matching algorithms are conducted by finding the disparity that can result in local minima for a pixel. Typical algorithms are cross-based local stereo matching with orthogonal integral images [4], local stereo matching with improved matching cost [5], etc. The regional minimization-based matching assumes that the disparities of pixels in a texture/color region can be modeled by a linear function. By using the initial disparities that are obtained from a local minimization-based matching, the disparities of the pixels in each region can be modified by regression [6]. The basic concept of global minimization-based matching is to estimate the disparities of all the pixels directly by modelling and solving a global energy function according to global optimization [7]. Typical global matching algorithms are dynamic programming-based algorithm [8], belief propagation algorithm [9][10], graph cut-based algorithm [11][12], and so on. There are different merits and demerits for each kind of algorithms. The complexity of the local minimization-based algorithms is very small, but the accuracy is somewhat not good enough. The global minimization-based algorithms usually give better results than the local minimization-based algorithms, but the complexity is very high. Besides, the boundary effect usually occurs because the global energy function cannot deal with the local texture and depth variations in the scene. The regional minimization-based algorithms seem to exploit the merits of local and global minimization-based algorithms simultaneously, but additional image segmentation is introduced, which increases the complexity.

Recently, because of the excellent performance in dealing with computer vision problems, convolutional neural network (CNN) has also been used for stereo matching [13]. The CNN-based stereo matching algorithms can also be divided into two kinds. One is the end-to-end training-based methods [14] [15], the other is to use CNNs to extract compact features of blocks in the left and the right images for further matching [16] [17] [18] [19].

In this paper, we propose an improved CNN-based local matching algorithm. Specifically, to efficiently extract the patch features, we add a deconvolution module in front of the network to learn how to enlarge the size of the input image patches. The de-convolved features of the left and the right view image patches are further fed into the successive convolutional layers with max-pooling to get compact features. Afterwards, these compact features are put together by dot product operation to generate a one-dimensional feature. Finally, softmax activation is performed based on the one-dimensional feature to estimate the disparity of the central pixel of the input left view image patch. Experimental results in the KITTI2012 and 2015 datasets show that more accurate disparities can be obtained by the proposed method.

The remainder of this paper is organized as follows. Related work is briefly introduced in Section II. The proposed algorithm is described in detail in Section III. Experimental results and conclusions are given in Section IV and V respectively.

**2.Related Work**

Traditional stereo matching algorithms mainly include SGM（Semi global matching）[20], BM(Block Matching) [21]，GC(Graph Cut) [22]，BP(Belief Propagation) [10] and DP (Dynamic Programming) [23]. In recent years, apart from the CNN-based stereo matching algorithms, the end-to-end method has also been developed well. N. Mayer et al. [24] created a large end-to-end deep neural network with the synthetic dataset, namely Dispnet. J. Pang et al. [25] proposed a two-stage coarse end-to-end model to refine the convolutional network for stereo matching. Recently, Khakis [26] proposed an end-to-end neural network that used the stacked hourglass backbone and increased the number of 3D convolutional layers for cost aggregation. However, the end-to-end method usually needs large memory and own large computational complexity, and is hard to converge [27].

For the CNN-based local matching algorithm, A. Geiger et al. [28] built a prior for the disparity to reduce the matching ambiguities of the remaining points. Akihito et al. [29] created two channels local window with conventional confidence features. The features and disparity patches were trained by CNN simultaneously. Chang et al. [15] proposed a pyramid stereo matching network, namely PSM-Net. The network structure consists of two main modules: spatial pyramid pooling and 3D CNN. It can fully exploit context information for finding correspondence in ill-posed regions. J. Zbontar, et al. [16] proposed a binary classification neural network with supervised training to calculate the matching error with small image patches. Xiao Liu et al. [17] proposed to use dilated convolution in the CNN to preserve the features of different scales so that better matching accuracy can be achieved. Compared to traditional convolution, the advantage of dilated convolution is that it can obtain a larger receptive field under the same convolution kernel size. However, too much dilated convolution will lead to too large a network structure and increase too much redundant information. Luo et al. [18] proposed a stereo matching algorithm by using the twin convolutional networks with shared weights in the left and the right network layers. In this method, the size of image patches ranges from 9x9 to 37x37. By using the constant size of the convolution kernel and the limited number of layers, the inference time for an image with resolution of 375x1275 is only 0.34s. To improve the matching accuracy of the boundary and the low-texture regions, Feng et al. [27] proposed a deeper neural network with a larger kernel size (such as 17x17), while the input images are up-sampled to two times.

In summary, to achieve higher accuracy, the patch size is usually enlarged. But larger patch size also needs to consume larger computational complexity. Moreover, the disparity map size decreases with the increase of the input patch size. For example, in the case that the original image size is 375x1242, it will be 347x1216, when using the network trained with an input patch size of 29x29. While the input patch size is 45x45, it will change into 331x1198. For the missing pixels at the edge of the image, it's disparity value is derived from the outermost layer of the disparity map. Besides, there is also an upper bound of the accuracy increment when increasing the input patch size. The above conclusions will be experimentally verified in Table7. In this paper, we expand the size of the image patch by deconvolution layers instead of using naive up-sampling methods and then get better results.

## 3.Proposed Method

We propose a type of CNN neural network with deconvolution layers [30] to adaptively expand the size of the input patches by referring to the GA-Net feature extraction. We introduce deconvolution at the beginning of the convolutional network and then perform the convolution operation. Deconvolution plays the role of upper sampling. The feature size after deconvolution is larger than the feature size before deconvolution. The backbone of the proposed network is based on the Siamese neural network [31] and MC-CNN network [19], we refer to the previous part of MC-CNN and combine the network architecture with the Siamese neural network. The basic structure of the proposed network is shown in Fig. 2. As can be seen, the overall network structure can be divided into four modules: image patch input module, deconvolution module, convolution module with dot product, and the Softmax-based loss calculation module. In the process of post-processing, we use the internal check method for checking the left and the right consistency, and use the ray filling method for filling invalid pixels. The flow chart of the overall algorithm is shown in Fig.3.

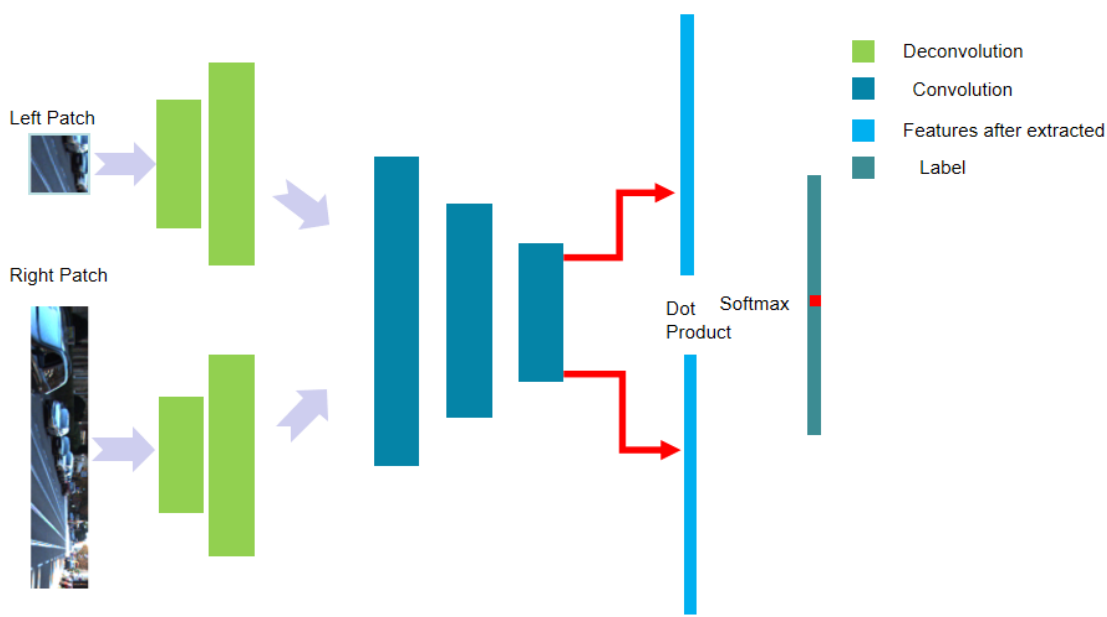

Fig. 2. ADSM network structure. This structure only consists of convolutional layer and deconvolution layer. At the end, it compares the label with dot product produced by the extracted features.

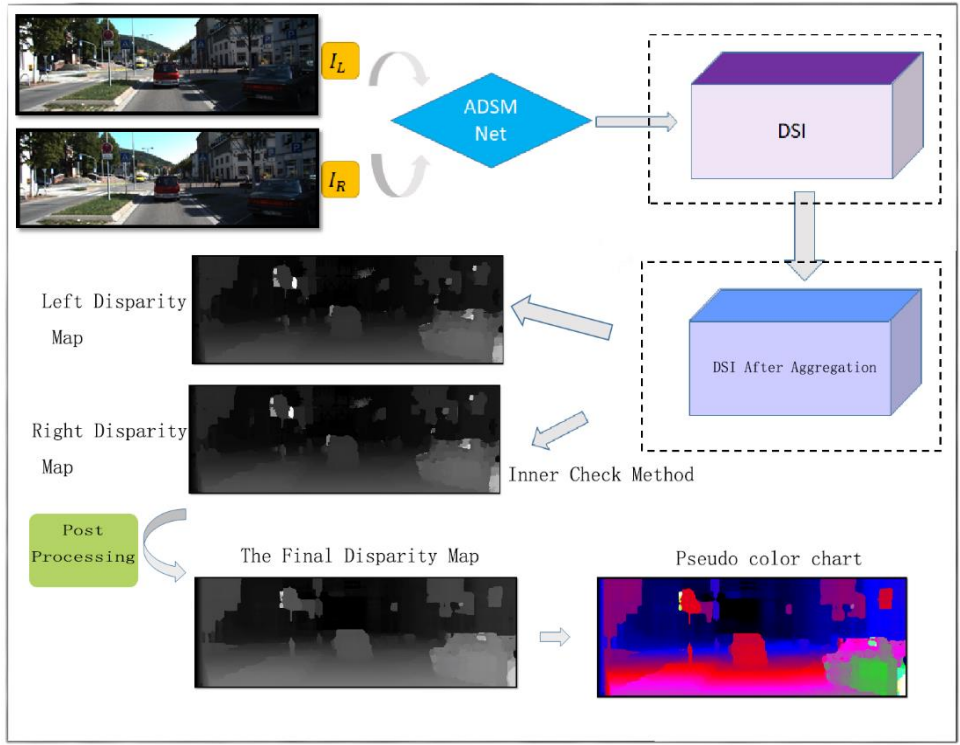

Fig3. The overall structure of the stereo matching algorithm. The left and right image will pass through ASDM Net and generate DSI. After 4-way cost aggregation, a new DSI will be generated. In this paper, we will use the quadratic fitting method to directly attend the sub-pixel level. After that, we use the inner check method to check the left and right consistency and finally, use the ray filling method to fill the invalid pixel position.

**Image patch input module**: Two image patches, denoted by $P_{W \times H}^{L}(p)$ and $P_{(W+200) \times H}^{R}(q)$ respectively, are input into the neural network, where the superscripts "$L$" and "$R$" denote the left and the right image patch, the subscripts "$W \times H$" and "$(W + 200) \times H$" represent the size of the two image patches, $p$ is the horizontal position of central pixel in the left image patch, and $q$ is the horizontal position of a pixel in the central line of the right image patch. The aim of the neural network is to determine whether the left image patch can be matched by a sub-patch with a size of $W \times H$ in the right image patch, as shown in Fig. 4.

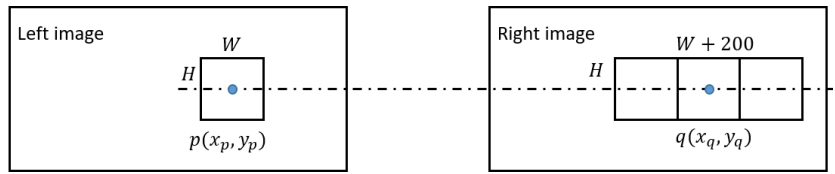

Fig. 4. Illustration of the input left and right image patches. Since the input images have been rectified by pole line, the center of the left and right image are all on the uniform line.

**Deconvolution module:**

| $i_{11}$ | $i_{12}$ |        | $h_{11}$ | $h_{12}$ |   | $o_{11}$ | $o_{12}$ | $o_{13}$ |
|----------|----------|--------|----------|----------|---|----------|----------|----------|
|          |          | $deconv$ |          |          | = | $o_{21}$ | $o_{22}$ | $o_{23}$ |
| $i_{21}$ | $i_{22}$ |        | $h_{21}$ | $h_{22}$ |   | $o_{31}$ | $o_{32}$ | $o_{33}$ |

Fig. 5. The operation of the one-step of deconvolution. Deconvolution is similar to convolution except that it can enlarge the original image.

As shown in Fig.5, let the input vector that is reshaped from a $2 \times 2$ matrix of the deconvolution operation be $i \in \mathbb{R}^{4\times 1}$, the output ($o \in \mathbb{R}^{9\times 1}$) vector that can be reshaped to a $3\times 3$ matrix of the deconvolution operation can be written as

$$o_{DC} = \mathcal{H}i, \tag{5}$$

where the subscript "DC" means deconvolution, $\mathcal{H} \in \mathbb{R}^{9\times 4}$ denotes the parameter matrix of the deconvolution operation,

$$\mathcal{H} = \begin{bmatrix} h_{11}, h_{12}, h_{21}, h_{22}, 0,0,0,0,0 \\ 0, h_{11}, h_{12}, 0, h_{21}, h_{22}, 0,0,0 \\ 0,0,0, h_{11}, h_{12}, 0, h_{21}, h_{22}, 0 \\ 0,0,0,0, h_{11}, h_{12}, 0, h_{21}, h_{22} \end{bmatrix}^{T}, \tag{6}$$

For the case of multi-deconvolution layers, the output of the deconvolution layer can be represented as

$$o_{DC} = \mathcal{H}_N\big(\cdots\big(\mathcal{H}_2(\mathcal{H}_1 i)\big)\big), \tag{7}$$

where, the $\mathcal{H}_1, ...,$ and $\mathcal{H}_N$ are parameters that should be learnt.

**Convolution module:** The aim of the convolution module is to extract compact features of the left and the right image patches for further matching. As we introduced in the deconvolution module, the input vector, the output vector, and the convolution operation can be expressed by $X \in \mathbb{R}^{9\times 1}$, $Y \in \mathbb{R}^{4\times 1}$, and $C \in \mathbb{R}^{4\times 9}$,

$$C = \begin{bmatrix} c_{11}, c_{12}, c_{21}, c_{22}, 0,0,0,0,0 \\ 0, c_{11}, c_{12}, 0, c_{21}, c_{22}, 0,0,0 \\ 0,0,0, c_{11}, c_{12}, 0, c_{21}, c_{22}, 0 \\ 0,0,0,0, c_{11}, c_{12}, 0, c_{21}, c_{22} \end{bmatrix}, \tag{8}$$

For the case of multi-convolutional layers, the output of convolution layer can be represented as

$$Y = C_N\big(\cdots\big(C_2(C_1 i)\big)\big), \tag{9}$$

where, the $C_1, ...,$ and $C_N$ are the parameters of each layer that should be learnt.

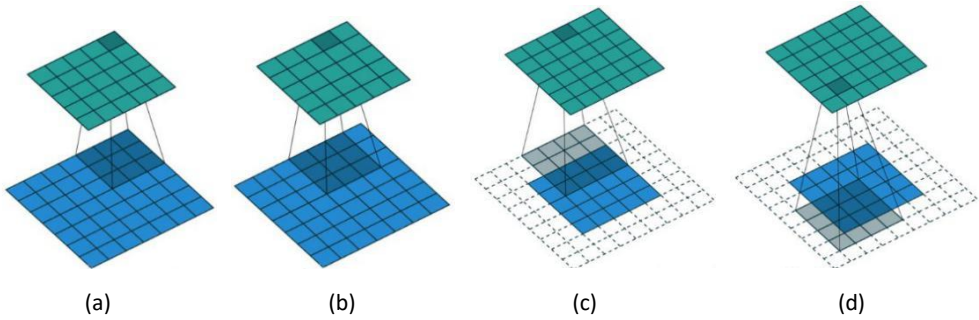

Fig.6. Illustration of convolution and deconvolution. (a), (b) represent the convolution operation, the green part represents the convolution result and blue part represents the input, the shadow part represents the size of convolution kernel. (c), (d) represent the deconvolution operation, the green part represents the deconvolution result, the blue part represents the input and the shadow part represents the kernel of deconvolution.

Fig. 6 shows the difference between convolution and deconvolution operations. The calculation relationship of the size of the convolutional network is

$$O = \frac{(I-k+2p)}{s} + 1, \tag{10}$$

while the calculation relationship of the size of the deconvolution network is

$$O = s(I - 1) - 2p + k,  \qquad (11)$$

where $O$ represents the output feature size, $I$ represents the input feature size, $s$ represents the step size, $p$ represents the fill size, and $k$ represents the convolution kernel size. In the convolutional neural network, the number of channels of the input feature is $CH_i$, the number of channels of the output feature map is $CH_o$, and the size of the convolution kernel is $w \times h$. Then the number of parameters of the convolutional layer is $CH_i \times h \times w \times CH_o + CH_o$. If Batch Normalization structure is adopted, $CH_o$[25] should be omitted.

As shown in Fig. 2, the left and the right image patch pass through their respective network branches. Each network branch contains different sizes of convolutional kernels. The size of the convolutional kernels of each layer varies according to the size of the previous layer. Each layer is then followed by a rectified linear units (ReLU) layer and batch normalization (BN) function except for the last layer. The output of the two branches are $\boldsymbol{O}_{left<1\times1\times64>}, \boldsymbol{O}_{right<1\times201\times64>}$ respectively, where

$$\boldsymbol{O}_{left} = [o_1^l, \cdots, o_{64}^l], \qquad (12)$$
$$\boldsymbol{O}_{right} = [\boldsymbol{o}_1^r, \cdots \boldsymbol{o}_n^r, \cdots, \boldsymbol{o}_{201}^r], \qquad (14)$$

and

$$\boldsymbol{o}_n^r = [o_1^r, \cdots, o_{64}^r], \qquad (15)$$

$n \in \{1 \dots 201\}$. Finally, the following dot product layer is adopted

$$\boldsymbol{r} = \boldsymbol{O}_{left}\boldsymbol{O}_{right}^T = \begin{bmatrix} o_1^l o_{1_1}^r + o_2^l o_{1_2}^r + \cdots + o_{64}^l o_{1_{64}}^r \\ \vdots \\ o_1^l o_{201_1}^r + o_2^l o_{201_2}^r + \cdots + o_{64}^l o_{201_{64}}^r \end{bmatrix}, \qquad (16)$$

to generate a vector $\boldsymbol{r}$ to indicate the matching degree of each possible disparities between the two image patches.

**Softmax-based loss calculation module**

In this module, we use Softmax classifier equipped by cross-entropy loss function to calculate the loss of each disparity. Note that there are 201 possible disparities in the proposed method. For each possible disparity $j \in \{1, \cdots 201\}$, the softmax function can be written to be

$$p_j = \boldsymbol{Softmax}(j) = \frac{e^j}{\sum_{i=1}^{201} e^i}, \qquad (17)$$

where $p_j$ denotes the probability of $j$. The outputs of all the possible disparities is then generated to be an output vector $\boldsymbol{p}_o = (p_1, \cdots p_{201})$. The cross-entropy loss can be calculated by

$$\boldsymbol{L}(\boldsymbol{\Theta}) = -\sum_{j=1}^{201} \boldsymbol{p}_{gt}(j) log(\boldsymbol{p}_o(j)), \qquad (18)$$

where $\boldsymbol{\Theta}$ denotes the current network parameters, $\boldsymbol{p}_{gt}$ is the ground truth label of each possible disparities. The ground truth label $\boldsymbol{p}_{gt}$ can be defined as a vector with only a single "1" element and "0" for all the other elements, where the position of the "1" element corresponds to the actual disparity, as shown in Fig. 7(a). To be more flexible, we define the $\boldsymbol{p}_{gt}$ to be

$$\boldsymbol{p}_{gt}(j) = \begin{cases} 0.5 & j = d_{gt} \\ 0.2 & |j - d_{gt}| = 1 \\ 0.05 & |j - d_{gt}| = 2 \\ 0 & otherwise, \end{cases} \qquad (19)$$

where $d_{gt}$ represents the groundtruth disparity value, as shown in Fig. 7(b). The training procedure is to find the optimal network parameter $\boldsymbol{\Theta}^{opt}$ to minimize the cross-entropy loss of (18).

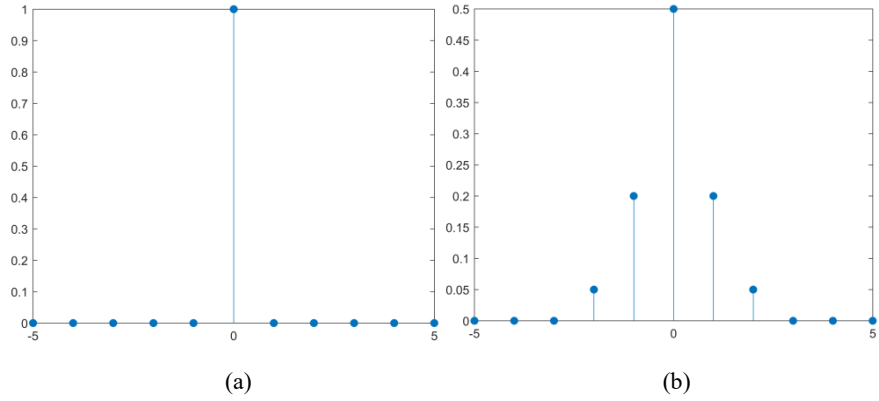

Fig.7 (a) and (b) represent different weight values for calculating cross entropy loss.

**Post Processing**

In the training procedure, we put the left and right images into the network model, the raw output of the neural network (whose softmax layer is excluded for test) is a three-dimensional array called DSI (Disparity Space Image). As shown in Fig.8(a), in which $\mathbb{H}, \mathbb{W}, \mathbb{D}$ represent the three dimensions of DSI respectively. Each element in the plane of $\mathbb{HW}$ represents the matching cost of the pixel under a certain disparity $d$. If the final disparities were selected based on the minimum matching cost of each pixel, noises will be inevitable, as shown in Fig. 9. Although by adding the deconvolution layers, noises can be reduced to some extent, the result is still not satisfactory. Therefore, we adopted the Semiglobal Matching (SGM) method [20] to further process the raw output of the convolutional network and generate a better disparity map.

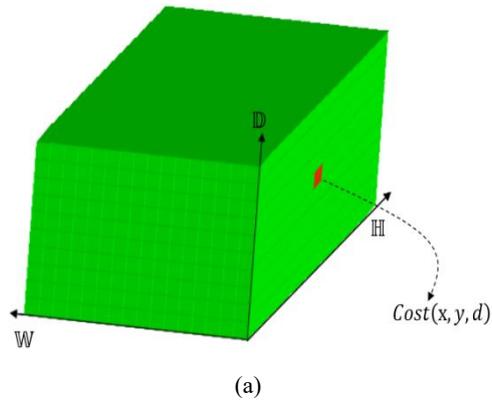

(a)

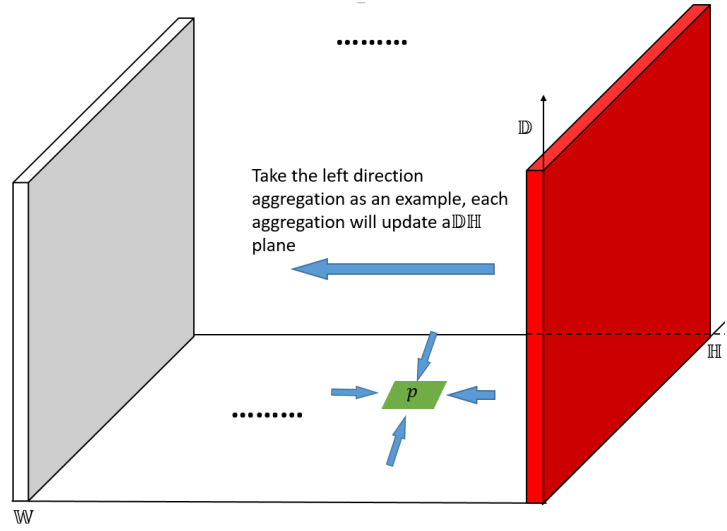

(b)

Fig.8 Disparity space image-based aggregation illustration, (a) represents disparity space image, (b) represents aggregation along with the left horizontal direction.

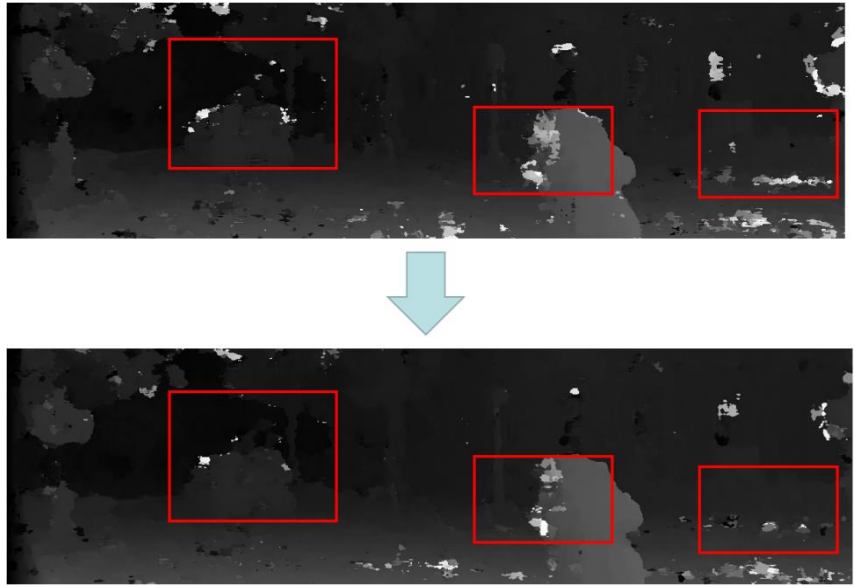

Fig.9 the upper image was generated by the proposed neural network with 4Conv, the following image was generated by 1Deconv(5)&4Conv. The input block size was set to be 37 x37.

Based on the SGM method, we select four directions (up, down, left, and right) for cost aggregation as shown in Fig.8(b). Take the left horizontal direction cost aggregation as an example, the cumulative cost iteration can be expressed as,

$$L_a(x,y,d) = Cost(x,y,d) + min\begin{cases} L_a(x-1,y,d), \\ L_a(x-1,y,d-1) + P_1, \\ L_a(x-1,y,d+1) + P_1, \\ min_d[L_a(x-1,y,d)] + P_2 \end{cases} - min_i\{L_a(x-1,y,d)\}, \quad (20)$$

where $(x,y,d)$ is the position of the cost cube, $P_1 = 30$ and $P_2 = 160$ are the predefined penalty parameters, $Cost(x,y,d)$ represents the matching cost obtained directly from the convolution network, and $L_a(x,y,d)$ is the cost after aggregation whose initial value is the output of the neural network, $min_d\{L_a(x-1,y,d)\}$ represents the minimum cost of the position $(x-1,y,d)$ in the cost cube for all possible disparity $d$. After the aggregation, one $\mathbb{HD}$ plane will be updated. Similarly, for the up and down directions, the corresponding $\mathbb{HW}$ plane will be updated. At last, the aggregate values of the four directions are added as the final DSI.

After the above steps, we still choose the disparity according to the principle of minimum cost value. To further improve the accuracy of the disparity map with low complexity, we adopt the inner check method [19] in the left/right consistency check step. The inner check method is to calculate the DSI of the right image through the DSI of the left image rather than network again. Let $Cost^L$ denote the DSI of the left image, $Cost^R$ denote the DSI of the right image, $D^L$ denote the disparity map obtained by $Cost^L$, and $D^R$ denotes the disparity map obtained by $Cost^R$. If $|D^L(x,y) - D^R(x-d,y)| \leq 1$, pixel $p(x,y)$ is valid in the $D^L$, otherwise we will mark it as an invalid pixel and assign it with disparity values from surrounding pixels.

## 4. Experimental results and analyses

We used KITTI2015 binocular dataset to verify the performance of the proposed neural network. The resolution of the images in the dataset is 1242x375. We first cropped each image into patches with size of 37x37 (for the left image) and 37x237(for the right image). 75% of the images were randomly selected for training, while the remaining images were used for test. During the training, the batch size is set to be 128. The NVIDIA GeForce GTX 1070Ti graphical card was used for training, and the iteration number was set to be 40000.

### 4.1 KITTI Datasets Description

KITTI stereo dataset is filmed by calibrated cameras while driving. The content of the dataset mainly includes roads, cities, residential areas, campuses, and pedestrians. The 3D laser scanner can obtain relatively dense distance information, and the depth map of the scene obtained by 3D laser scan can be used as the real disparity map after calculation. The KITTI platform provides 3D point cloud data obtained by 3D scanning lasers, corresponding calibration information, and coordinate transformation information, from which we can generate the real disparity maps.

### 4.2 Patch pair generation

We first cropped the left image into patches with a size of 37x37. Then, the ground truth disparity of the central pixel $p$ of the patch is recorded. For each of the cropped left image patch, we found the pixel position $q$ in the right image according to the ground truth disparity of the pixel $p$. Centered by the position $q$, we cropped the patch with the size of 37x237 as the input right image patch, as shown in Fig. 3. Then the input of the proposed neural network can be denoted as $<\boldsymbol{P}^L_{37\times37}(p), \boldsymbol{P}^R_{37\times237}(q)>$. Therefore, for the position $p$ in the left image patch, there are 200 potential matching positions,

$$\{P^R_{37\times237}(q-100), \ldots P^R_{37\times237}(q+100)\},$$

and the aim is of the neural network is to select the optimal matching position. To this end, we also define the label of the neural network to be $\{0,\cdots,0.05,0.2,0.5,0.2,0.05,\cdots,0\}$, as shown in (19) and Fig. 7.

### 4.3 Image sample processing

In the KITTI2015 dataset, the disparity map is saved in "png" format and the data type is uint16. In the disparity map, the point with a gray value of 0 is an invalid point, that is, there is no real disparity value at this point [12]. For these points, we did not put them into the training set. The disparity map processing is expressed as follow:

$$DispValue(x,y) = ((float)Disp(x,y))/256.0 \qquad (20)$$

where, $Disp(x,y)$ represents the original disparity map, $DispValue(x,y)$ represents the processed disparity map. The image patch should be further normalized to [0, 1] for the input of the proposed neural network.

**4.4 Training**

We first tested six configurations of layers and convolutional kernels for the proposed neural network without deconvolution layers, as shown in Table 1. 3Conv in the table represents 3 convolution layer, Conv13 represents 13x13 the size of the convolution kernel in the convolution layer, and so forth. By taking the deconvolution layers into the network structure, the overall convolution neural network configurations are given in Table 2, in which 1Deconv (3) stands for a single deconvolution layer with a kernel of 3x3.

Besides, we also verified the effectiveness of the proposed method with different image block input sizes. We try to set the size of the left image block to 29 x29, 33 x33, 37 x37, 41 x41, and 45 x45. Accordingly, the right input image block sizes were set to be 29x229, 33x233, 37x237, 41x241, 45x245.

Table1. Six configurations of convolutional layers for the proposed neural network.

| 3Conv | 4Conv | 6Conv |
|---|---|---|
| Conv13+BN+ReLU | Conv10+BN+ReLU | Conv9+BN+ReLU |
| Conv13+BN+ReLU | Conv10+BN+ReLU | Conv9+BN+ReLU |
| Conv13+ softmax | Conv10+BN+ReLU | Conv7+BN+ReLU |
|  | Conv10+softmax | Conv7+BN+ReLU |
|  |  | Conv5+BN+ReLU |
|  |  | Conv5+softmax |

| 7Conv | 9Conv | 11Conv |
|---|---|---|
| Conv7+BN+ReLU | Conv5+BN+ReLU | Conv5+BN+ReLU |
| Conv7+BN+ReLU | Conv5+BN+ReLU | Conv5+BN+ReLU |
| Conv7+BN+ReLU | Conv5+BN+ReLU | Conv5+BN+ReLU |
| Conv7+BN+ReLU | Conv5+BN+ReLU | Conv5+BN+ReLU |
| Conv5+BN+ReLU | Conv5+BN+ReLU | Conv5+BN+ReLU |
| Conv5+BN+ReLU | Conv5+BN+ReLU | Conv5+BN+ReLU |
| Conv5+softmax | Conv5+BN+ReLU | Conv5+BN+ReLU |
|  | Conv5+BN+ReLU | Conv3+BN+ReLU |
|  | Conv5+softmax | Conv3+BN+ReLU |
|  |  | Conv3+BN+ReLU |
|  |  | Conv3+softmax |

Table2. The overall convolution neural network configurations for the proposed neural network.

| 1Deconv(5)&4Conv | 1Deconv(3)&4Conv | 2Deconv&6Conv | 3Deconv&6Conv |
|---|---|---|---|
| Deconv5+BN | Deconv3+BN | Deconv3+BN | Deconv3+BN+ReLU |
| Conv11+BN+ReLU | Conv11+BN+ReLU | Deconv5+BN | Deconv5+BN+ReLU |
| Conv11+BN+ReLU | Conv11+BN+ReLU | Conv9+BN+ReLU | Deconv7+BN+ReLU |
| Conv11+BN+ReLU | Conv10+BN+ReLU | Conv9+BN+ReLU | Conv9+BN+ReLU |
| Conv11+softmax | Conv10+softmax | Conv9+BN+ReLU | Conv9+BN+ReLU |
|  |  | Conv7+BN+ReLU | Conv9+BN+ReLU |
|  |  | Conv7+BN+ReLU | Conv7+BN+ReLU |
|  |  | Conv7+softmax | Conv7+BN+ReLU |

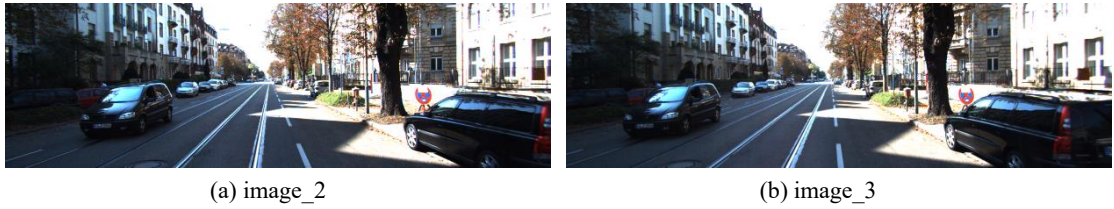

(a) image_2　　　　　　　　　　　　　　　(b) image_3

Fig.10. Sample images of KITTI2015 dataset.

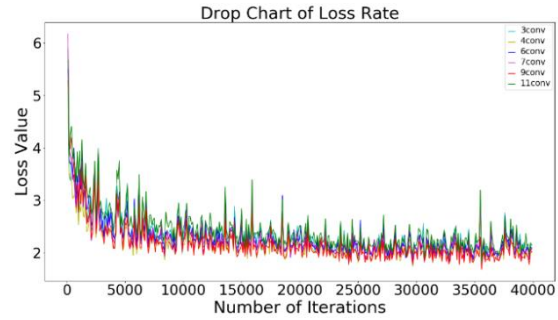

Fig. 11. Loss curve of different network configurations.

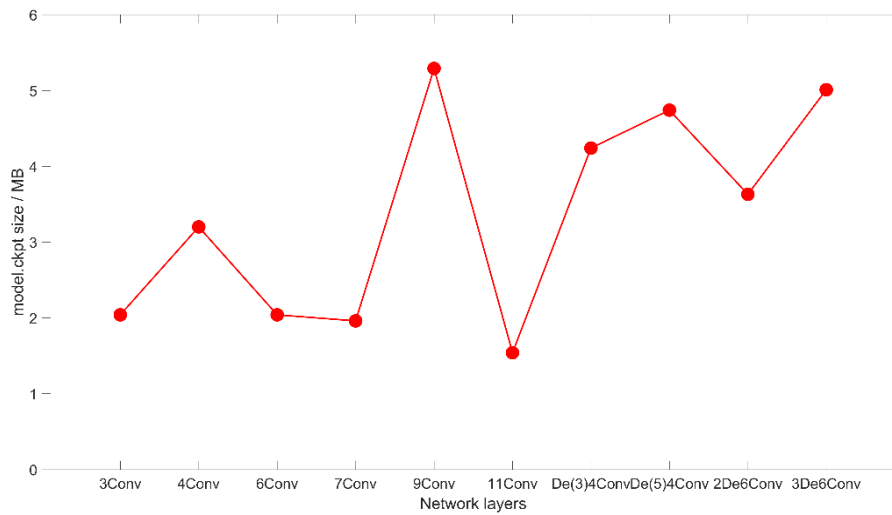

Fig.12 The network model (model.ckpt file) size after training for different network structures. The ckpt file is a binary file from tensoflow, which stores all the weights, bias, gradients and other variables.

Table 3. The percentage of missing matching pixels with threshold of 2, 3, 4, 5 in the 2015 KITTI test set.

| ConvNet-ValError | 2-pixel-error | 3-pixel-error | 4-pixel-error | 5-pixel-error |
|---|---|---|---|---|
| 37-3Conv | 12.21 | 8.55 | 6.95 | 6.05 |
| 37-4Conv | 11.11 | 8.12 | **6.82** | **6.01** |
| 37-6Conv | 11.79 | 8.73 | 8.35 | 6.53 |
| 37-7Conv | 11.96 | 8.87 | 7.53 | 6.73 |
| 37-9Conv | **10.86** | **8.07** | **6.82** | 6.07 |
| 37-11Conv | 14.95 | 10.36 | 8.82 | 7.92 |

Table 4. The number of network parameters of different network structures for input left patch size 37.

| NetworkType | 3Conv | 4Conv | 6Conv | 7Conv | 9Conv | 11Conv |
|---|---|---|---|---|---|---|
| Number of unilateral network parameters | 1416896 | 1248000 | 953536 | 918720 | 721600 | 627932 |

Table 5. The number of network parameters with deconvolution structure for input left patch size 37.

| NetworkType | 1Deconv(5)&4Conv | 1Deconv(3)&4Conv | 2Deconv&6Conv | 3Deconv&6Conv |
|---|---|---|---|---|
| Number of unilateral network parameters | 1987264 | 1812160 | 1701568 | 1902182 |

Table6. The results with deconvolution structure and other network for input left patch size 37.

| ConvNet-ValError | 2-pixel-error | 3-pixel-error | 4-pixel-error | 5-pixel-error |
|---|---|---|---|---|
| 37-1Deconv(5)&4Conv | **10.31** | **7.60** | **6.38** | **5.74** |
| 37-1Deconv(3)&4Conv | 10.36 | 7.67 | 6.49 | 5.78 |
| 37-2Deconv&6Conv | 13.63 | 10.56 | 9.27 | 8.53 |
| 37-3Deconv&6Conv | 15.65 | 12.46 | 11.17 | 10.45 |
| MC-CNN-acrt[19] | 15.02 | 12.99 | 12.04 | 11.38 |
| MC-CNN-fast[19] | 18.47 | 14.96 | 13.18 | 12.02 |
| Efficient-Net[18] | 11.67 | 8.97 | 7.62 | 6.78 |

Table7. The percentage of missing matching pixels with threshold of 2, 3, 4, 5 in the 2015 KITTI test set for input left patch size 29, 33, 41, 45.

| Input Patch Size | ConvNetModel | 2-pixel-error | 3-pixel-error | 4-pixel-error | 5-pixel-error |
|---|---|---|---|---|---|
| 29×29 | 3Conv | 13.49 | 10.29 | 8.79 | 7.92 |
| | 4Conv | 12.76 | 9.98 | 8.63 | 7.84 |
| | 5Conv | 12.8 | 10.1 | 8.78 | 8.03 |
| | 7Conv | 13.21 | 10.44 | 9.13 | 8.36 |
| | 1Deconv(3)-4Conv | **12.47** | **9.89** | **8.71** | **8.02** |
| | 1Deconv(2)-4Conv | 12.96 | 10.28 | 9.04 | 8.32 |
| 33×33 | 4Conv | 12.34 | **9.59** | **8.25** | **7.45** |
| | 8Conv | 13.11 | 10.31 | 8.63 | 7.84 |
| | 1Deconv(3)-4Conv | **12.20** | 9.70 | 8.55 | 7.88 |
| | 1Deconv(5)-4Conv | 12.80 | 10.20 | 9.03 | 8.35 |
| 41×41 | 4Conv | 10.61 | 7.69 | 6.41 | 5.66 |
| | 5Conv | 10.39 | 7.64 | 6.43 | 5.69 |
| | 8Conv | 11.03 | 8.21 | 6.96 | 6.21 |
| | 10Conv | 11.47 | 8.51 | 7.21 | 6.45 |
| | 1Deconv(3)-4Conv | **9.94** | **7.31** | **6.15** | **5.47** |
| | 1Deconv(5)-4Conv | 10.26 | 7.51 | 6.29 | 5.56 |
| 45×45 | 4Conv | 10.48 | 7.55 | 6.24 | 5.47 |
| | 5Conv | 10.54 | 7.69 | 6.44 | 5.69 |
| | 8Conv | 10.78 | 7.88 | 6.62 | 5.85 |
| | 1Deconv(3)-4Conv | **9.45** | **6.86** | **5.72** | **5.03** |

Table8. The percentage of missing matching pixels with threshold of 2, 3, 4, 5 in the 2012 KITTI validation set.

| ConvNetModel | 2-pixel-error | 3-pixel-error | 4-pixel-error | 5-pixel-error |
|---|---|---|---|---|
| MC-CNN-acrt[21] | 16.92 | 14.93 | 13.98 | 13.32 |
| MC-CNN-fast[21] | 19.56 | 17.41 | 16.31 | 15.51 |
| Efficient-Net[15] | 12.86 | 10.64 | 9.65 | 9.03 |
| Ours | **8.42** | **7.07** | **6.43** | **6.02** |

Table9. Error comparison of disparity with different algorithms (KITTI2012) %.

| Algorithm | 2-pixel-error | 3-pixel-error | 4-pixel-error | 5-pixel-error | Runtime(s) |
|---|---|---|---|---|---|
| StereoSLIC[33] | 7.20 | 5.11 | 4.04 | 3.33 | 2.3 |
| PCBP-SS[32] | 6.75 | 4.72 | 3.75 | 3.15 | 300 |
| SPSS[32] | 6.28 | 4.41 | 3.52 | 3.00 | 2 |
| MC-CNN-acrt[19] | 5.45 | 3.63 | 2.85 | 2.39 | 67 |
| Displets v2[34] | **4.46** | **3.09** | **2.52** | **2.17** | 265 |
| Efficient-CNN[18] | 6.51 | 4.29 | 3.36 | 2.82 | **0.7** |
| Ours | 5.62 | 4.01 | 3.02 | 2.65 | 2.4 |

Table10. Error comparison of disparity with different algorithms (KITTI2015) %.

| Algorithm | 2-pixel-error | 3-pixel-error | 4-pixel-error | 5-pixel-error | Runtime(s) |
|---|---|---|---|---|---|
| Elas[28] | 24.09 | 19.21 | 17.59 | 16.82 | 0.669 |
| SGM[20] | 10.03 | 6.93 | 5.47 | 4.48 | 1.8 |
| SPSS[32] | 7.15 | 4.58 | 3.46 | 2.93 | 3 |
| Efficient-CNN[18] | 6.78 | 4.38 | 2.56 | 2.03 | 1 |
| MC-CNN-fast[19] | 7.53 | 4.01 | 2.84 | 2.33 | **0.2** |
| MC-CNN-slow[19] | 6.38 | **3.27** | **2.37** | **1.97** | 35 |
| Ours | **4.27** | 3.85 | 2.57 | 2.00 | 2.5 |

## 4.6 Comparison

Fig.11 shows the loss curve of each network configuration. We can see that the network with 9 convolutional layers and 4 convolutional layers converges better than the other configurations. Table 3 shows the percentage of missing matching pixels with a threshold of 2, 3, 4, 5 in the test set, in which 37×37 represents the size of the input image block. The threshold means the absolute difference between the estimated disparity and the actual disparity. We can see that the network with 9 convolutional layers configuration performs better for 2, and 3 pixels errors (with percentages of 10.86% and 8.07%), whereas, the network with 4 convolution layers configuration performs better for 5 pixels error (with a percentage of 6.01%). For the threshold of 4, the performances of them are the same as each other. (the percentage is 6.82%).

Besides, Table 4 compares the number of parameters of different network configurations. We can see that the more convolutional layers produce fewer network parameters. In Fig.12, we can see that the network structure which takes up more space of "ckpt" file can get better matching results. But it is difficult to obtain a good matching result if the network parameters are too small such as the network with 11Conv configuration.

Table 6 shows the performances of the neural network with different deconvolution layers. we can see the 1Deconv(5) with 4Conv configuration owns the best performance in the table. The corresponding 2, 3, 4, and 5 pixels errors are only 10.31%, 7.60%, 6.38%, and 5.74%, whereas the errors are 15.02%, 12.99%, 12.04%, and 11.38% for the method in Zbontar[19]. We can also see that excessive usage of deconvolution layers will not always improve the matching accuracy. We believe the reason is that although deconvolution can strengthen the edge information and equip them better model expression ability, more deconvolution layers will introduce more irrelevant elements such as "0" in formula 6.

Table 5 compares the number of parameters of different network configurations with deconvolution structure. The size of image layers will be increased because of the operation of the deconvolution kernel. Therefore, in the following, the size of the convolution kernels will also be enlarged correspondingly. From Tables 4 and 5, we can see that the best matching result can be obtained with a constant convolution kernel size and appropriate deconvolution structure layers such as 1Deconv(5)&4Conv.

As shown in Table 7, we can see that the proposed neural network with deconvolution layers can achieve better matching results in different image block sizes comparing to the neural network without deconvolution layers. When the input image block size is 29x29, the best result of the neural network without deconvolution layers (4Conv configuration) is 12.76%, 9.98%, 8.63%, 7.84% for the 2, 3, 4, and 5 pixels errors. As a comparison, the best result of the neural network with deconvolution layers (1Deconv(3)-4Conv configuration) is 12.47%, 9.89%, 8.71%, and 8.02%. That is to say, benefit from the deconvolution layers, better results can be achieved for the 2 and 3 pixels errors. Furthermore, we also enlarged the input image block size to discuss the influence of the block size. When the input image block size is 41x41, the best result of the neural network without deconvolution layers (5Conv configuration) is 10.39%, 7.64%, 6.43%, and 5.69% for the 2, 3, 4, and 5 pixels errors, while that with deconvolution layers (also 1Deconv(3)-4Conv configuration) is 9.94%，7.31%，6.15%，5.47%, which are all better than those without deconvolution layers. In summary, equipped with the deconvolution layers, better results can be achieved in most cases, indicating the effectiveness of the proposed method.

To further verify the effectiveness of the proposed neural network, we also did the experiments with the KITTI 2012 validation set. As shown in Table 8, better results can also be achieved by the proposed neural network. The corresponding 2, 3, 4, and 5 pixels errors are only 8.42%, 7.07%, 6.43%, 6.02%, whereas the errors are 12.86%, 10.64%, 9.65%, and 9.03% for the method in [18].

In addition, we also compared the proposed method with the other state-of-the-art methods, MC-CNN[19], Efficient-CNN[18], Elas[28], SGM[20], SPSS[32], PCBP-SS[32],StereoSLIC[33], Displets v2[34] in Tables 9 and 10, respectively. From Table 9, we can see that, for the KITTI2012 dataset, the DispletsV2[34] is the best, the corresponding 2, 3, 4, and 5 pixels errors are 4.46%, 3.09%,

2.52%, and 2.17%. But we should also note that the accuracy is achieved by great computational complexity. The processing time for a picture on average is 265 seconds. The MC-CNN-acrt also achieves good results, i.e., the 2, 3, 4, and 5 pixels errors are 5.45%, 3.63%, 2.85%, and 2.39%. But its processing time is still large, i.e., 67 seconds. The proposed method can achieve comparable accuracy (i.e., 5.62%, 4.01%, 3.02%, and 2.65% for 2,3,4, and 5 pixels errors) with a much small processing time, i.e., 2.4 seconds. For the results of the KITTI2015 dataset, as shown in Table 10, we can see that the proposed method can achieve the smallest 2 pixels error (4.27%). The 3, 4, and 5 pixels errors (3.85%, 2.57%, and 2.00%) of the proposed method is only a little larger than MC-CNN-slow which achieves the best accuracy on the whole. However, the processing time of the proposed method is only 2.5 seconds on average, which is much smaller than MC-CNN-slow.

**5.Conclusion and future work**

For improving the accuracy of stereo matching, based on CNN stereo matching, we designed a novel neural network structure by adding a deconvolution operation before the formal convolutional layers. The introduced deconvolution operation at the beginning of the network improves the size of the input features maps for the following convolution layers, as a results the whole convolution network learns more information. Besides, we also dropped the fully connected layer and replaced it with a dot product to enable it to obtain matching results in a short time. Experimental results demonstrate that better matching results can be achieved by the proposed neural network structure (with the configuration of 37-1Deconv(5)&4Conv) with low inference and training time complexity.

6.Acknowledgment

This work was supported in part by the Shenzhen Science and Technology Research and Development Funds (Grants No.JCYJ20170818104011781) and in part by the National Natural Science Foundation of China under Grants 61871342. The authors would like to thank the editors and anonymous reviewers for their valuable comments.

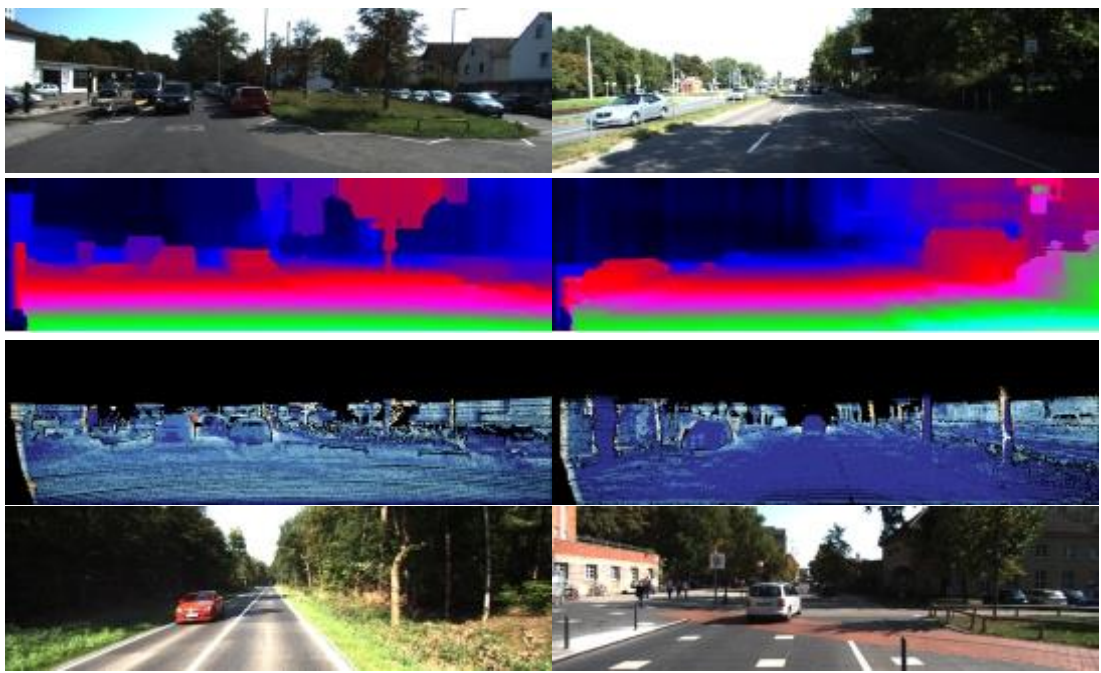

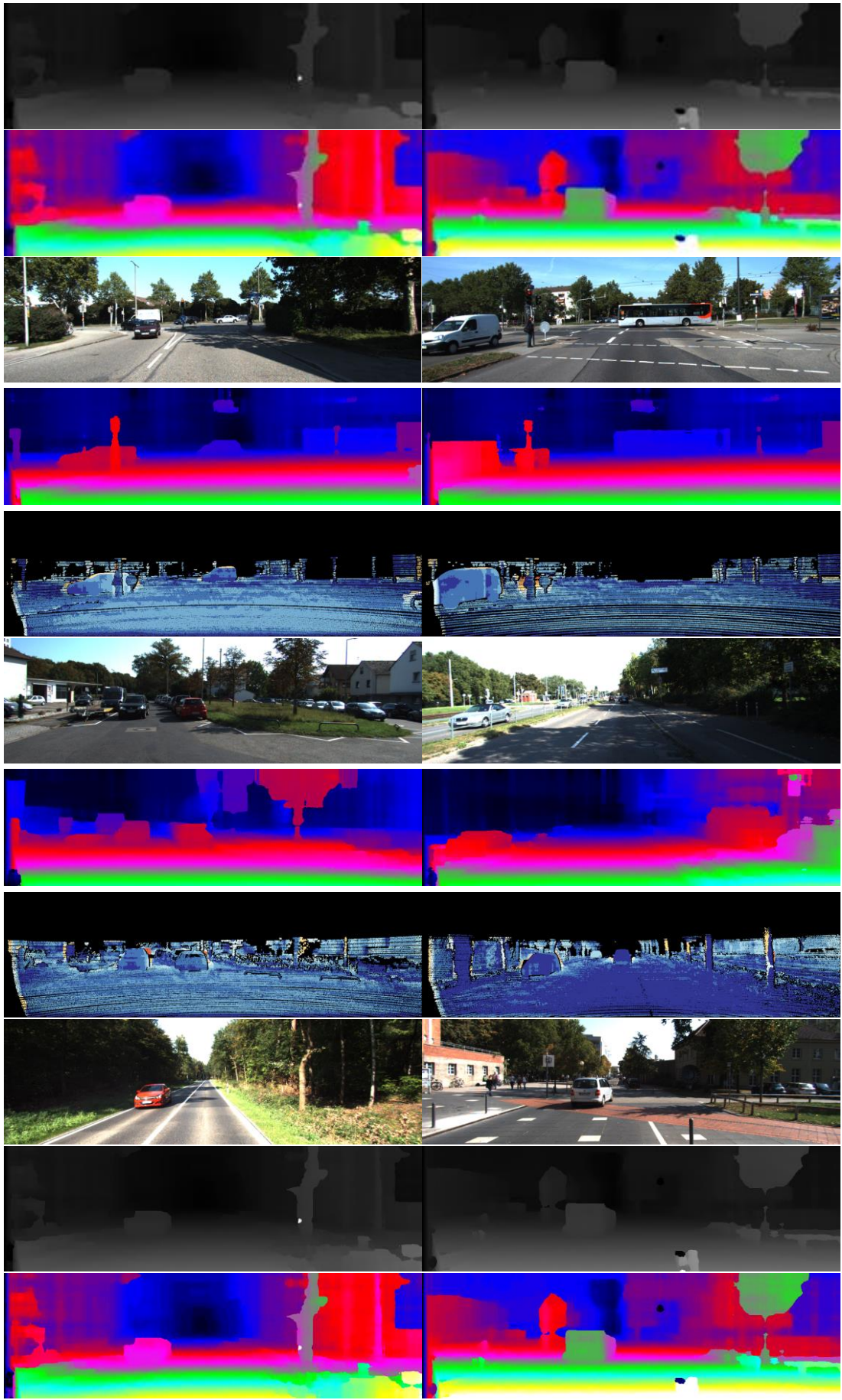

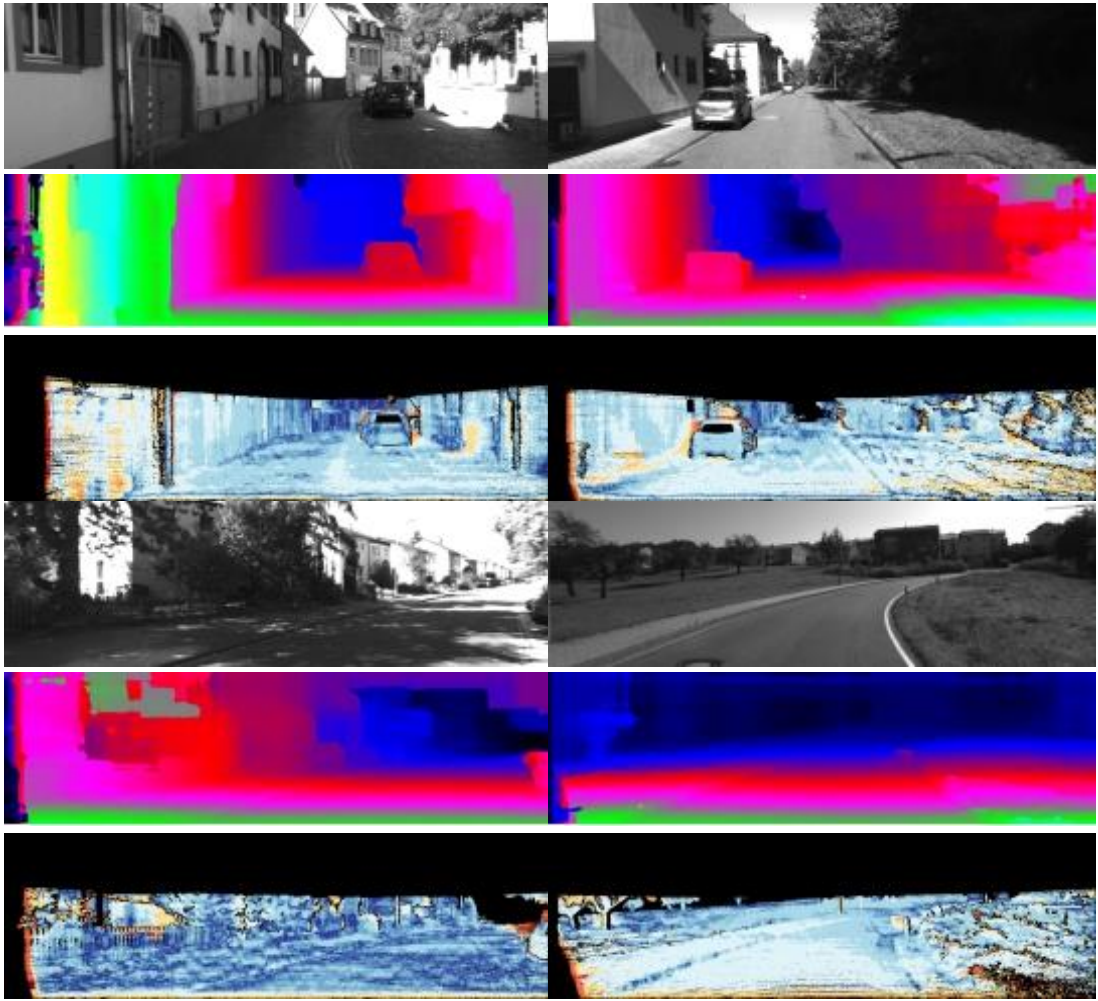

Fig.12 KITTI2015 test set: (left) original image, (center) stereo estimates, (right) stereo errors.

Fig.13 KITTI2015 test set: (left) original image, (center) stereo estimates, (right) stereo errors.